\documentclass{article}
\usepackage[T1]{fontenc} 
\usepackage[utf8]{inputenc} 
\usepackage{ismir,amsmath,cite,url}
\usepackage{graphicx}
\usepackage{color}

\usepackage{soul} 
\usepackage{multirow}
\usepackage{mathtools}
\usepackage{dsfont}
\usepackage{color}
\usepackage{hyperref}

\usepackage{lineno}

\title{Data Cleansing with Contrastive Learning for Vocal Note Event Annotations}

\multauthor
{Gabriel Meseguer-Brocal$^1$ \hspace{1cm} Rachel Bittner$^2$ \hspace{1cm} Simon Durand$^2$ \hspace{1cm} Brian Brost$^2$} {
$^1$ STMS UMR9912, Ircam/CNRS/SU, Paris. $^2$ Spotify, USA.\\
{\tt\small gabriel.meseguerbrocal@ircam.fr, \{rachelbittner, durand, brianbrost\}@spotify.com}
}
\def\authorname{Gabriel Meseguer-Brocal, Rachel Bittner, Simon Durand, Brian Brost}

\begin{document}

\maketitle
\begin{abstract}
Data cleansing is a well studied strategy for cleaning erroneous labels in datasets, which has not yet been widely adopted in Music Information Retrieval.
Previously proposed data cleansing models do not consider structured (e.g. time varying) labels, such as those common to music data.
We propose a novel data cleansing model for time-varying, structured labels which exploits the local structure of the labels, and demonstrate its usefulness for vocal note event annotations in music.
We frame the problem as an instance of contrastive learning, where we train a model to predict if an audio-annotation pair is a match or not.
We generate training data for this model by automatically deforming known correct annotations to form incorrect annotations.
We demonstrate that the accuracy of a transcription model improves greatly when trained using our proposed strategy compared with the accuracy when trained using the original dataset.
Additionally we use our model to estimate the annotation error rates in the DALI dataset, and highlight other potential uses for this type of model.
\end{abstract}

\section{Introduction}
Labeled data is necessary for training and evaluating supervised models, but the process of creating labeled data is often error prone.
Labels may be created by human experts, by multiple human non-experts (e.g. via crowd sourcing), semi-automatically, or fully automatically.
For many problem settings, even in the best case scenario where data is labeled manually by experts, labels will almost inevitably have inconsistencies and errors.
The presence of label noise is problematic both for training and for evaluation \cite{Frenay_2014}. During training, it can cause models to converge slower and to require much more data, or overfit the noise thus resulting in poor generalization.
During evaluation, it can lead to unreliable metrics with artificially low scores for models with good generalization and artificially high scores for models which overfit noisy data. 
This issue is particularly timely and relevant for the music information retrieval (MIR) community as recent datasets such as LakhMIDI~\cite{Raffel_2016}, DALI~\cite{Meseguer-Brocal_2018} or the Free Music Archive~\cite{Deferrard_2017} take advantage of large music collections accessible from the Internet but often rely on noisy annotations. Additionally, many common annotation tasks are particularly costly as they have to be aligned in time and require a participant with musical expertise to be done accurately.

\emph{Data cleansing} is a well studied technique in the machine learning community for mitigating the effects of label noise, with a focus on improving model generalization when training on noisy datasets~\cite{Frenay_2014}.
A common and effective approach is to build a model to identify and discard data points with incorrect labels. Most methods taking this approach do not assume any structure or correlation between different labels. This is appropriate for many common tasks, such as image recognition. However, in music, labels are often highly structured and time-varying, and the label noise is not random.
For example, musical note-annotations, which we focus on in this work, are locally stable in time and follow certain common patterns.
Typical noise for note events include incorrect pitch values, shifted start times, and incorrect durations, among others.

Our contributions are as follows. We propose a novel contrastive learning-based~\cite{wu2018unsupervised, chen2020simple} data cleansing model which can exploit sequential dependencies between labels to predict incorrectly labeled time-frames trained using likely correct labels pairs as positive examples and and local deformations of correct pairs as negative examples.
We focus our experiments on a model for detecting errors in vocal note event annotations, which we believe extends easily to other types of music transcription labels.
We then demonstrate the usefulness of this data cleansing approach by training a transcription model on the original and cleaned versions of the DALI~\cite{Meseguer-Brocal_2018} dataset.
Further, we use the model to estimate the error rates in the DALI dataset, and highlight other potential uses for this type of model, including for reducing manual labeling efforts.
Finally, the code used in this work, including the pre-trained error detection model, is made freely available~\footnote{https://github.com/gabolsgabs/contrastive-data-cleansing} along with the outputs of the model for the DALI dataset~\footnote{https://zenodo.org/record/3576083}.

\section{Background and Related Work}

We first introduce prior work on learning in the presence of label noise, and conclude by summarizing the work relevant to our example use case of note event annotations.

\subsection{Classification in the presence of label noise}

We consider the problem of training a classifier on a dataset where some of the labels are incorrect.
One class of solutions attempts to solve the problem by mitigating the effect of label noise, rather than modifying the data used for training. This is commonly done by modifying the loss function to directly model the distribution of label noise, for example, by creating a noise-robust loss with an additional softmax layer to predict correct labels during training~\cite{Goldberger_2017}, or with a generalized cross-entropy that discards predictions that are not confident enough while training, looking at convergence time and test accuracy~\cite{Zhang_2018}, or by inferring the probability of each class being corrupted into another~\cite{Patrini_2017}.
However, these approaches are restricted to specific types of loss functions or make restrictive assumptions about the statistical distribution of noise.



\textbf{Data cleansing} \cite{Brodley_1999} and \textbf{outlier detection} \cite{Guo_2018} based approaches aim to identify the correctly labeled data points and train only on them. The vast majority of data cleansing methods are model prediction-based \cite{Frenay_2014}.
In their simplest form, model prediction-based methods train a model to remove items from the dataset where the label predicted by the model disagrees with the dataset label. Note that in many cases, the choice of model for data cleansing is often the same as the choice of model used after data cleansing.



These data cleansing approaches have several advantages over learning directly with noisy labels.
First, the filtering does not depend on the downstream inference task, thus a cleansing method can be applied to filter data used to train many different models. Second, we can train less complex downstream models, as they do not need to account for label noise. To the best of our knowledge however, prior data cleansing approaches do not exploit the structured nature of labels often seen in MIR tasks.

In addition to developing data cleansing methods, or learning methods that are robust to label noise, there are a variety of less closely related paradigms for dealing with data quality issues. In \textbf{semi-supervised learning} reliably labeled data is combined with a large amount of unlabeled data~\cite{Zhu_2005}. In
\textbf{weakly supervised learning}~\cite{Mintz_2009, Mnih_2012, Xiao_2015} low-quality or insufficiently granular labels are used to infer the desired target information. Finally, \textbf{active learning} estimates the most valuable unlabeled points for which to solicit additional labels~\cite{Settles_2008, Krause_2016}.

\subsection{Note Event Annotations}

Automatic music transcription, one of the core tasks in MIR, involves converting acoustic music signals into some form of music notation~\cite{benetos2019automatic}. \emph{Musical note events} are a common intermediate representation, where a note event consists of a start time, end time and pitch.
They are useful for a number of applications that bridge between the audio and symbolic domain, including symbolic music generation and melodic similarity.
Instruments such as the piano produce relatively well-defined note events, where each key press defines the start of a note.
Other instruments, such as the singing voice, produce more abstract note events, where the time boundaries are often related with changes in lyrics or simply as a function of our perception~\cite{furniss2016michele}, and are therefore harder to annotate correctly.

Datasets providing note event annotations are created in a variety of ways, all of which are error prone.
Notes may be manually labeled by music experts, requiring the annotator to specify the start time, end time and pitch of every note event manually, aided by software such as Tony~\cite{mauch2015computer}.
MIDI files from the Internet can in some cases be aligned automatically as in the LakhMIDI~\cite{Raffel_2016} and DALI~\cite{Meseguer-Brocal_2018} datasets, with varying degrees of accuracy and completeness.
Note data has also been collected automatically using instruments which ``record'' notes while being played, such as a Disklavier piano in the MAPS~\cite{emiya2009multipitch} and MAESTRO~\cite{hawthorne2018enabling} datasets, or a hexaphonic guitar in the GuitarSet dataset~\cite{xi2018guitarset}.
Data collected in this way is typically quite accurate, but may suffer from global alignment issues~\cite{hawthorne2018enabling} and can only be achieved for these special types of instruments.
Another approach is to play a MIDI keyboard in time with a musical recording, and use the played MIDI events as note annotations~\cite{su2015escaping} but this requires a highly skilled player to create accurate annotations.

\begin{figure}[th]
 \centerline{
   \includegraphics[width=0.45\textwidth]{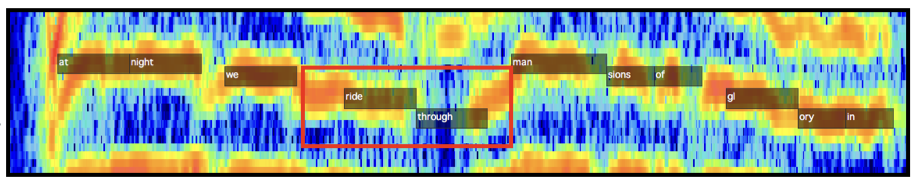}
 }
 \caption{Example of two incorrect note annotations from the DALI dataset, outlined in red, figure from~\cite{meseguerbrocal_2020}.}
 \label{fig:wrongexample}
\end{figure}

\figref{fig:wrongexample} shows an example of correct and incorrect note annotations in the DALI dataset.
The types of errors produced by the previous methods can vary.
A single note can be imprecise in time, resulting in an incorrect start time or duration, or the pitch value can be annotated wrong.
Additionally, notes can be annotated where there are no actual notes in the audio, and conversely, notes in the audio can be missed all together.
Systematic errors include global shifts and stretches in time and shifts in key/octave.
Local errors are difficult to detect, and systematic errors can cause every note event to be wrong in some way.
At the individual time-frame level, notes with incorrect start/end times will have errors at the beginning/ending frames, but can still be correct in the central frames.

\section{Data Cleansing for Note Events}\label{sec:our_approach}

Given an input space $\mathcal{X}$ and a label space $\mathcal{Y}$, we propose a model prediction based approach, but rather than training a classifier $h: \mathcal{X} \rightarrow \mathcal{Y}$, we directly train a model $g: (\mathcal{X} \times \mathcal{Y}) \rightarrow [0, 1]$ which approximates the probability that the label is incorrect. We denote this probability by $g(x, \hat y)$. 

Note that this is mathematically equivalent in the ideal case to the previous model prediction based approaches: Given a perfect estimator $h$ which always predicts a correct label $y$, $g(x,y)=\mathds{1}_{h(x)\ne y}$ where $\mathds{1}$ is the indicator function.
However, for complex classification tasks with high numbers of classes and structured labels, modeling $h$ can be much more complex than modeling $g$.
For instance, consider the complexity of a system for automatic speech recognition, versus the complexity needed to estimate if a predicted word-speech pair is incorrect. Intuitively, you don't need to know the right answer to know if something is right or wrong.

This idea is similar to the ``look listen and learn''~\cite{arandjelovic_2017} concept of predicting the ``correspondence'' between video frames and short audio clips -- two types of structured data.
It is also similar to CleanNet~\cite{Lee_2017cleannet}, where a dedicated model predicts if the label of an image is right or wrong by comparing its features with a class embedding vector. However, this approach operates on global, rather than position-dependent labels. 

\begin{figure*}
    \centering
    \includegraphics[width=1.5\columnwidth]{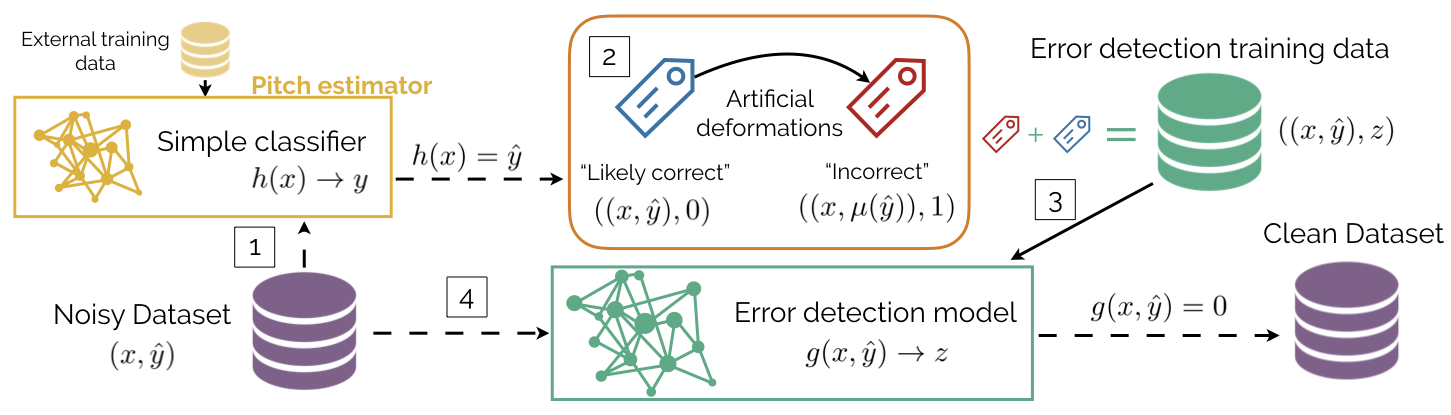}
    \caption{Overview of our data cleansing system. The training data for our error detection model $g$ is generated automatically. 1- We predict the $f_0$ for the whole noisy dataset and compare it with the annotated label. 2- We select as ``likely correct'' examples those where the prediction is similar to the label, distorting them to generate the incorrect examples. 3- We train our $g$ using this new training set. 4- We filter the noisy dataset obtaining the clean version.}
    \label{fig:sys_diagram}
\end{figure*}

In our approach, we generate training data for $g$ by directly taking pairs $(x,y)$ from the original dataset as positive examples and creating artificial distortions of $y$ to generate negative examples.
In this section, we study the use of an estimator $g(x, \hat{y})$ for detecting local errors in noisy note-event annotations.
See \figref{fig:sys_diagram} for an overview of the system.

\subsection{Input Representations}\label{sec:inout}
As our input representation, instead of using the raw audio signal itself, we compute the Constant-Q Transform (CQT)~\cite{Brown_1991} as a matrix $X$, where $X_{ij}$ is a time-frequency bin.
The time index $i$ corresponds to the time stamp $r_i = \upsilon \cdot i$ where $\upsilon$ is a constant defining the spacing between time stamps, and the frequency index $j$ corresponds to a frequency $q_j$ in Hz.
The CQT is a bank of filters transformation centered at geometrically spaced frequencies.
We use a frequency bin resolution with 6 octaves, 1 bin per semitone, a sample rate of $22050$ Hz and a hop size of $256$, resulting in a time resolution of $\upsilon = 11.6$ ms.
We compute the CQT from the original mixture and from the isolated vocal version derived from the mixture using a source separation technique~\cite{Jansson_2017}.
We include the CQT of the isolated vocals to boost the information in the signal related to the singing voice, and couple it with the CQT of the mixture to include information we may have lost in the separation process.

We define the annotated label $\hat Y$ as a binary matrix created from the original note-event annotations. For a given track, let $K$ be the set of note annotations, let $t_k^0$ and $t_k^1$ be the start and end time of note $k$ in seconds, and $f_k$ be its frequency in Hz. Then $\hat Y$ is defined as:
\begin{equation} \label{eq:dali_labels}
\begin{small}
  \hat{Y}_{ij} =
   \begin{cases}
      1, & \text{ if } t_k^0 \le r_i \le t_k^1, \, q_{j-1} < f_k \le q_{j}, \, k \in K \\
      0, & \text{ otherwise}
  \end{cases}
\end{small}
\end{equation}
and has the same time and frequency resolution as $X$.

\subsection{Learning Setup}
We estimate the label noise at the \textbf{\emph{time frame}} level. 
Let $\ell \in L$ be an index over the set of tracks $L$, $I^{\ell}$ be the index of all time frames for track $\ell$, and $X^{\ell}$ and $\hat Y^{\ell}$ be its CQT and label matrices respectively.
Let $X^{\ell}_i$ and $\hat Y^{\ell}_i$ indicate a time frame of $X^{\ell}$ and $\hat Y^{\ell}$
that contains all its frequency bins $j$,
so we index the data points according to their time index only.
Finally, let $X^{\ell}_{a:b}$ and $\hat Y^{\ell}_{a:b}$ denote the sequence of time frames of $X$ and $\hat Y$ between time indices $a$ and $b$.


Our goal is to identify the subset of time frames $i \in I^{\ell}$ which have errors in their annotation for each track in a dataset by training a binary data cleansing model.
Our data cleansing model is a simple estimator that can be seen as a binary supervised classification problem that produces an error detection model.
Given a datapoint centered at time index $i$, $g$ predicts the likelihood that the label $\hat Y_i$ is wrong.
Critically, we take advantage of the structured labels (i.e. the temporal context); as input to $g$ we use $X_{a:b}$ and $Y_{a:b}$ to predict if the center frame $\hat Y_{(a+b)/2}$ of $\hat Y_{a:b}$ is incorrect.
That is, we aim to learn $g$ such that:
\begin{equation} \label{eq:cleansing}
  g(X_{a:b}, \hat Y_{a:b}) = \begin{cases}
      0, & \text{ if } \hat{Y}_{(a+b)/2} \text{ is correct}  \\
      1, & \text{ if } \hat{Y}_{(a+b)/2} \text{ is incorrect}
  \end{cases}
\end{equation}
Thus, in order to evaluate if a label $\hat Y_i$ is correct using $n$ frames of context, we can compute $g(X_{i-n:i+n}, \hat Y_{i-n:i+n})$. In the remainder of this work, we will define
\begin{equation}
    g_n(X_i, Y_i) := g(X_{i-n:i+n}, \hat Y_{i-n:i+n})
\end{equation}
as a shorthand.
In this work, we use $n=40$.

Let
\begin{equation}\label{eq:dali_set}
D = \bigcup_{\ell \in L} I^{\ell}
\end{equation}
be the set of all time indices of all tracks in $L$. Our aim is to use $g$ to create a filtered index $F$, where:

\begin{equation}\label{eq:filtered_index}
F = \{i \in D : g_n(X_i, \hat Y_i) = 0 \}
\end{equation}

\subsection{Training data generation}\label{sec:quality_data}
Let $z_i$ be a binary label indicating whether the center frame $\hat Y_i$ for an input/output pair $(X_{i-n:i+n}, \hat Y_{i-n:i+n})$ is incorrect.
To train $g$, we need to generate examples of correct and incorrect data-label pairs $\left((X_{i-n:i+n}, \hat Y_{i-n:i+n}), z_i\right)$.
We will again introduce a shorthand $((X_i, Y_i), z_i)$ to refer to data points of the form  $\left((X_{i-n:i+n}, \hat Y_{i-n:i+n}), z_i\right)$.


\textbf{Incorrect Data} To generate ``incorrect'' data points (where $z_i = 1$), we can simply randomly distort any of the existing labels in the dataset by applying a modification function $\mu(\hat Y_i)$.
These modifications $\mu(\hat Y_i)$ are not random but rather specific to match the characteristics of typical note-event errors (issues in the positions of the start or end times, incorrect frequencies, or the incorrect absence/presence of a note).
These distorted $\hat Y_i$ should be contextually realistic, meaning that notes should have a realistic duration and should not overlap with the previous or next note.
To do this, we modify the original note events $(t_k^0, t_k^1, f_k)$ described in Section~\ref{sec:inout} by randomly shifting start and end times, frequency values, and by randomly deleting or adding notes.
Given these new note events, we generate a new label matrix $\mu(\hat Y_i)$ as described in~\eqnref{eq:dali_labels}.
Some note-level modifications may still result in the center frame being ``correct'' -- for example, if a note begins a few frames late, the following frames will still be correct -- thus, after modifying creating $\mu(\hat Y)$, we sample examples from frames where the center frame $\mu(\hat Y_i) \ne \hat Y_i$.

\textbf{Likely Correct Data} Since we do not have direct access to the true label $Y_i$ (indeed this is what we aim to discover), we first use a ``simple classifier'' proxy for selecting likely correct data points.
There are many possible choices for this proxy -- for example, if there is a manually verified subset of a dataset, it can be used directly as the set of likely correct data points -- in this work we outline one specific example.
We first compute the output of a pre-trained $f_0$ estimation model $s(X_i)$ that given $X_i$ outputs a matrix with the likelihood that each frequency bin contains a note~\cite{Bittner_2017}.
$s(X_i)$ is trained on a different dataset than we use in the subsequent experiments and has been proven to achieve state-of-the-art results for this task~\cite{Bittner_2017}.
$s(X_i)$ produces $f_0$ sequences, rather than note events, which vary much more in time than note events, so we define an agreement function in order to determine when the labels agree.
$s(X_i)$ is not a perfect classifier, and while its predictions are not always correct, we have observed that when the agreement is high, $\hat Y_i$ is usually correct.  However, we cannot use low agreement to find incorrect examples, because there are many cases with low agreement even though $\hat{Y_i}$ is correct.
Therefore, we only use $\kappa$ to select a subset of ``likely correct'' data points.

We compute both ``local'' (single-frame) and ``patch-level'' (multi-frame) agreement, and use thresholds on both to select time frames which are likely correct.
The local agreement, $\kappa_l$ is computed as:
\begin{equation}
    \kappa_l(\hat Y_i, s(X_i)) = \max_j \left( \hat Y_{ij} \cdot s(X_i)_{j} \right)
\end{equation}
and the patch-level agreement $\kappa_p$ is a $k$ point moving average over time of $\kappa_l$.
For the test set of $g$, we use very strict thresholds and select $(X_i, \hat Y_i)$ pair to be a likely correct if $\kappa_{l} > .999$ and $\kappa_{p} > .85$.
For the training set, we use more relaxed thresholds, and select points with $0.9 < \kappa_l \le 0.999$ and $0.7 < \kappa_p \le 0.85$.
These values have been found manually and assure the selection of good ``likely correct'' examples.
This procedure gives us a set of positive examples in non-silent regions, but does not take into account the silent areas.
In order to select correctly labeled points from silent regions, we take additional  $(X_i, \hat Y_i)$ points from regions with low energy in the isolated vocals and no annotations in a window of length $v$. In this work we use $v = 200$ ($\approx 2,32$ s).

Finally, the combination of ``likely correct'' and ``incorrect'' data results in a dataset of $\{((X_i, \hat Y_i), z_i) \}$ with which we can train $g$.

\subsection{Error Detection Model Architecture}\label{sec:model}
\begin{figure*}[ht]
 \centerline{
   \includegraphics[width=.8\textwidth]{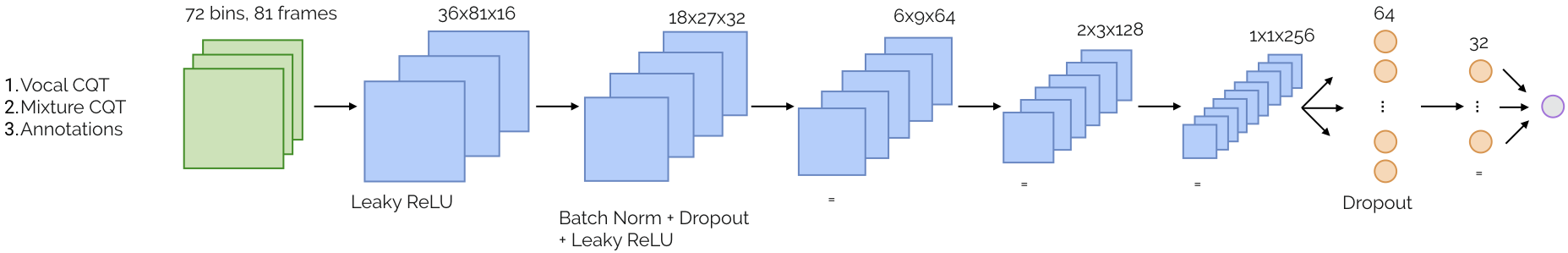}
 }
 \caption{Error detection model architecture.}
 \label{fig:quality_model}
\end{figure*}

We propose a standard convolutional architecture for our error detection model $g_n(X_i, \hat Y_i) = z_i$, as shown in \figref{fig:quality_model}.
The input of the model is a matrix with $72$ frequency bins, $81$ time frames ($0.94$ seconds) and three channels: the two CQTs (mixture and vocals) $\{X_{i-n:i+n}\}$ and the label matrix $\{\hat Y_{i-n:i+n}\}$.
It has five convolutional blocks with $3 \times 3$ kernels, `same` mode convolutions, with leaky ReLU activations for the first block and batch normalization, dropout and leaky ReLU for the rest. The strides are [$(2, 1)$, $(2, 3)$, $(3, 3)$, $(3, 3)$, $(2, 3)$] and the number of filters [$16$, $32$, $64$, $128$, $256$] generating features maps of dimensions  $(36\times 81\times 16)$, $(18\times 27\times 32)$, $(6\times 9 \times 64)$, $(2\times 3\times 128)$, $(1\times 1\times 256)$.
Then, we have two fully-connected layers with $64$ and $32$ neurons, a ReLU activation and dropout and a last fully-connected layer with one neuron and a sigmoid activation.
The model is trained using a binary cross entropy loss function.

\section{Experiments}
 \label{sec:cleansing_validation}

We test our approach using the DALI dataset, version 2~\cite{meseguerbrocal_2020}, which contains 7756 user-submitted vocal MIDI annotations from karaoke websites, automatically matched and aligned to polyphonic audio files.
The MIDI annotations are crowd sourced and thus have highly varying quality, while the alignment is done automatically and likely to contain mistakes. This dataset is then particularly relevant for this work.
The contrastive learning-based error detection model $g$ is trained as described in Section~\ref{sec:model} using the data generation method described in Section~\ref{sec:quality_data}.
The trained model $g$ had a frame-level accuracy of 72.1\% on the holdout set, and 76.8\% on the training set.

\subsection{Training with Cleaned Data}

\begin{figure}[ht]
 \centerline{
   \includegraphics[width=.42\textwidth]{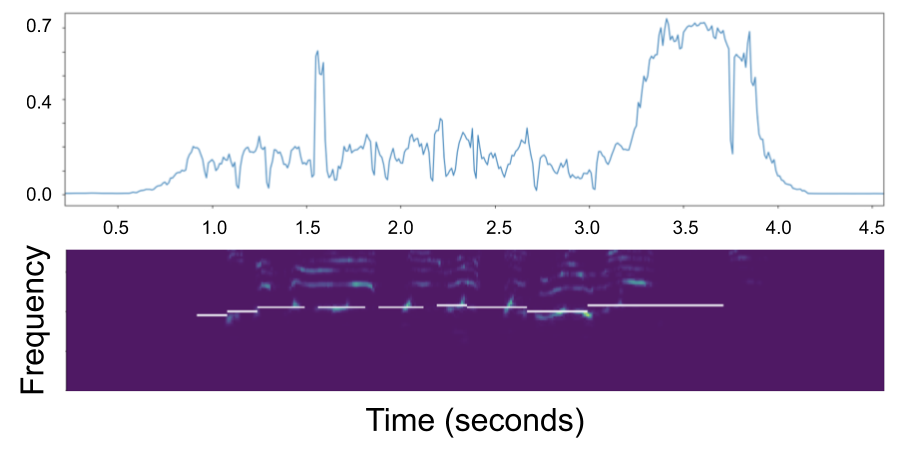}
 }
 \caption{(Top) The output of the error detection model for a short segment. (Bottom) the corresponding CQT and annotated notes (in white). The error is high at the beginning of the fourth note because it starts late, and at end of the last note because it is too long.}
 \label{fig:quality_example}
\end{figure}

Validating the performance of $g$ is challenging, as we only have likely correct and artificially created wrong examples, but we do not have any ``real'' ground truth correct and incorrect examples.
Thus, we first manually verified the predictions of the error detection model in many random examples (see~\figref{fig:quality_example}), and found that they appeared to be strongly correlated with errors in $y_i$.
However, a manual perceptual evaluation of the error detection model is both infeasible and defeats the purpose of automating the process of correcting errors. Instead, we validate the usefulness of this approach by applying it to model training.
In this section, we address the question: \textbf{Is the error detection model useful? How much?}

The ultimate goal of a data cleansing technique such as this one is to identify incorrect labels and remove them from the dataset in order to better train a classifier.
Thus, one way to demonstrate the effectiveness of the data cleansing method is to see if training a model using the filtered dataset results in better generalization than training on the full dataset.

To validate the usefulness of $g_n(X_i, \hat Y_i)$ for improving training, we train the Deep Salience vocal pitch model~\cite{Bittner_2017} three times\footnote{We train each new model from scratch, not using transfer learning} using three different training sets.
The training sets are subsets of DALI, and contain the $\hat{Y_i}$ of all the songs that have a Normalized Cross-Correlation $> .9$~\cite{Meseguer-Brocal_2018}.
This results in a training set of 1837 songs. The three sets are defined as follows:

\begin{enumerate}
    \item \textbf{All} data. Trained using all time frames, $D$ (\eqnref{eq:dali_set}).
    \item \textbf{Filtered} data. Trained using the filtered, ``non-error'' time frames, $F$ (\eqnref{eq:filtered_index}), where the output of $g$ has been binarized with a threshold of 0.5. With this data we tell the model to skip all the estimated noisy labels.
    \item \textbf{Weighted} data.  Trained using all time frames, $D$, but during training, the loss for each sample is weighted by $1 - g_n(X_i, \hat Y_i)$. This scales the contribution of each data point in the loss function according to how likely it is to be correct.
\end{enumerate}

\begin{figure}[ht]
 \centerline{
   \includegraphics[width=.46\textwidth]{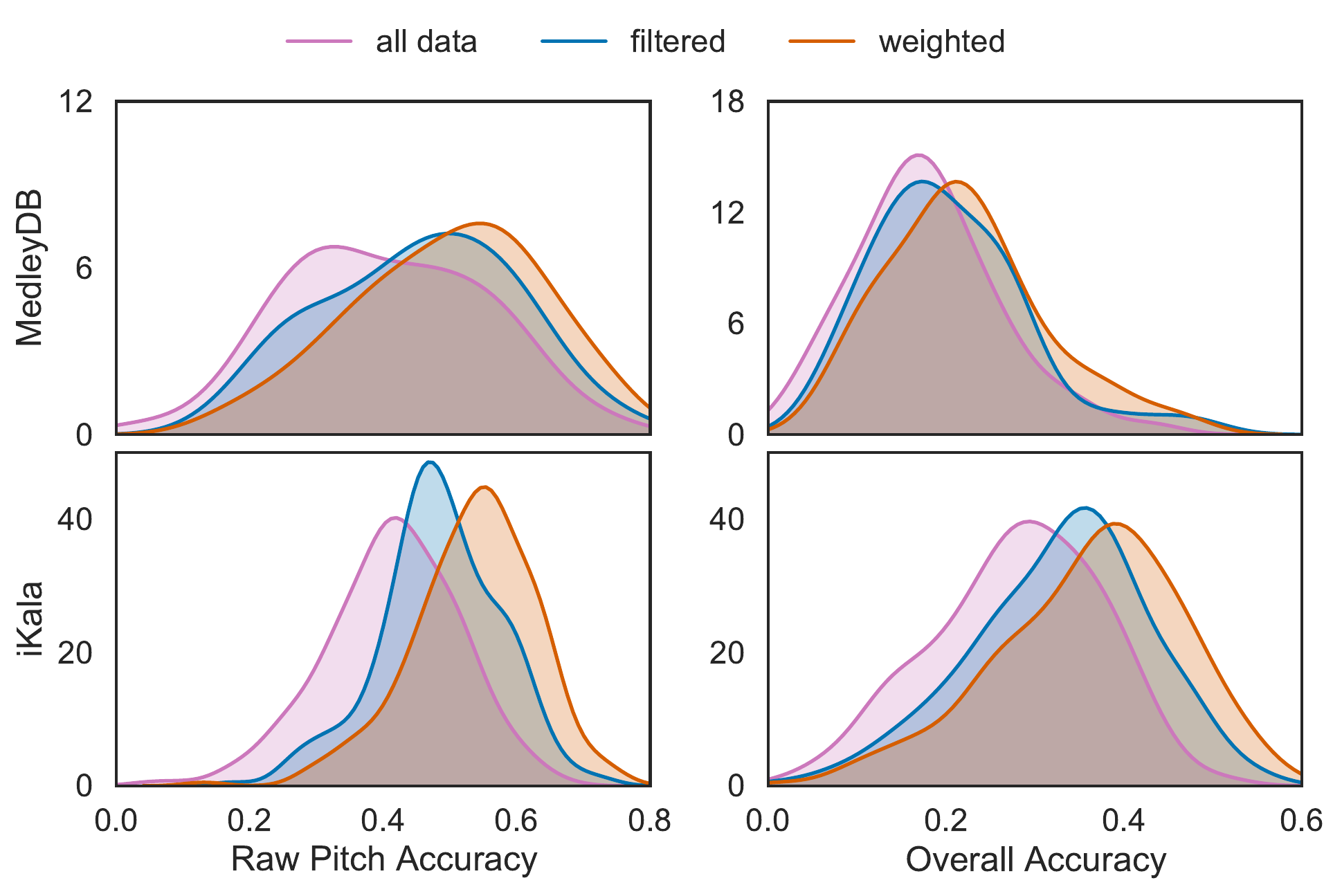}
 }
 \caption{Distribution of scores for the three training conditions. Each condition is plotted in a different color. Scores for MedleyDB are shown in row 1 and scores for iKala are in row 2. Raw pitch accuracy is shown in column 1 and overall accuracy is shown in column 2. The y-axis in all plots indicates the number of tracks.}
 \label{fig:quality_exp_hist}
\end{figure}

We test the performance of each model on two polyphonic music datasets that contain vocal fundamental frequency annotations: 61 full tracks with vocal annotations from MedleyDB~\cite{Bittner_2014} and 252 30-second excerpts from iKala~\cite{Chan_2015}.
We compute the the generalized\footnote{using continuous voicing from the model output and binary voicing from the annotations} Overall Accuracy (OA) which measures the percentage of correctly estimated frames, and the Raw Pitch Accuracy (RPA) which measures the percentage of correctly estimated frames where a pitch is present, which are standard metrics for this task~\cite{bittner2019,salamon2014melody}.
The distribution of scores for each dataset and metric are shown in \figref{fig:quality_exp_hist}.

While the scale of the results are below the current state of the art~\cite{Bittner_2017} -- likely due to the noisiness of the training data! -- we see a clear positive impact on performance when data cleansing is applied.
The overall trend we see is that training using filtered data outperforms the baseline of training using all the data with statistical significance ($p < 0.001$ in a paired t-test) for all cases, indicating that our error detection model is successfully removing time frames which are detrimental to the model.
We also see that, overall, training using all the data but using the error detection model to weigh samples according to their likelihood of being correct is even more beneficial than simply filtering.
This suggests that the likelihoods produced by our error detection model are well-correlated with the occurrence of real errors in the data.
These trends are more prominent for the iKala dataset than for the MedleyDB dataset -- in particular, the difference between training on filtered vs. weighted data is statistically insignificant for MedleyDB while it is statistically significant ($p < 0.001$ in a paired t-test) for the iKala dataset.
The iKala dataset has much higher proportion of \emph{voiced} frames (frames with a pitch annotation) than MedleyDB. This suggests that the weighted data is beneficial for improving pitch accuracy, but may not bring any improvement over filtering for detecting whether a frame should have a pitch or not (voicing).
Nevertheless, both conditions which used the error detection model to aid the training process see consistently improved results compared with the baseline.


\subsection{Estimated Quality of The DALI Dataset}

\begin{figure}
    \centering
    \includegraphics[width=.85\columnwidth]{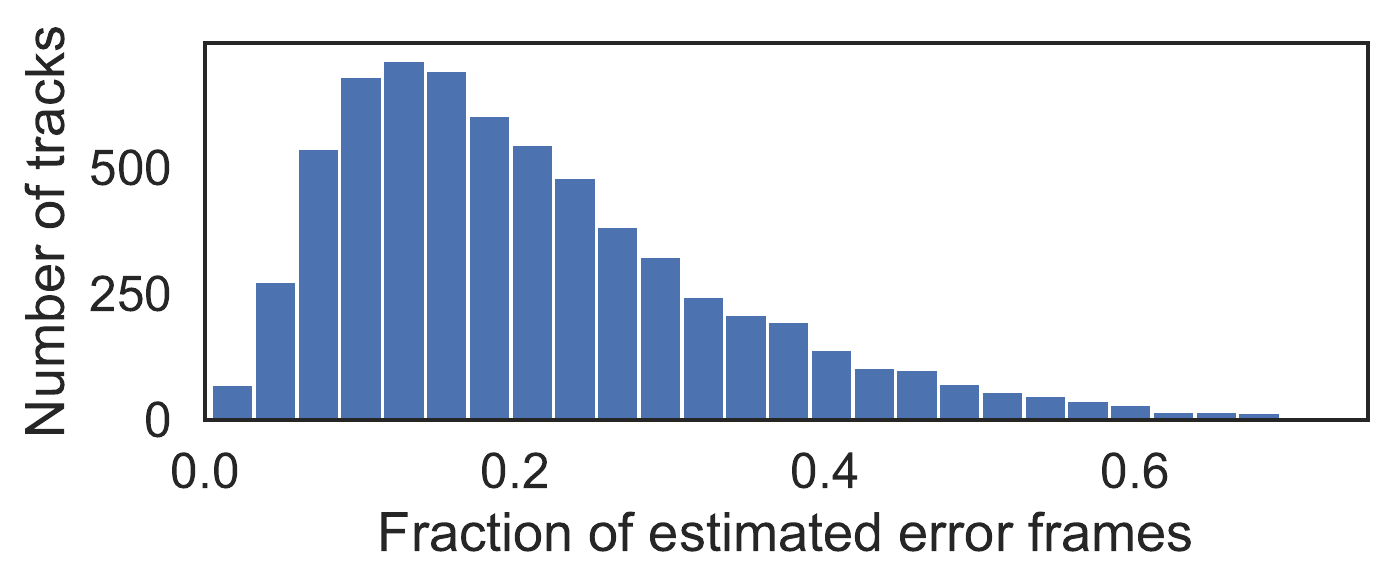}
    \caption{Histogram of the estimated error rate per track.}
    \label{fig:quality_hist}
\end{figure}

As a final experiment, we ran the error detection model on the full DALI dataset (version 2) in order to estimate the prevalence of errors.
We compute the percentage of frames per-track where the likelihood of being an error is $\ge 0.5$.
A histogram of the results is shown in \figref{fig:quality_hist}.
We estimated that on average, 21.3\% of the frames of a track in DALI will have an error in the note annotation, with a standard deviation of 12.7\%.
31.1\% of tracks in DALI have more than 25\% errors, while 18.2\% of tracks have less than 10\% errors.
We also measure the relationship between the percentage of estimated errors per track and the normalized cross correlation from the original DALI dataset~\cite{Meseguer-Brocal_2018}, and found no clear correlation.
This indicates that while the normalized cross correlation is a useful indication of the global alignment, it does not reliably capture the prevalence of local errors.

\begin{figure}
    \centering
    \includegraphics[width=.9\columnwidth]{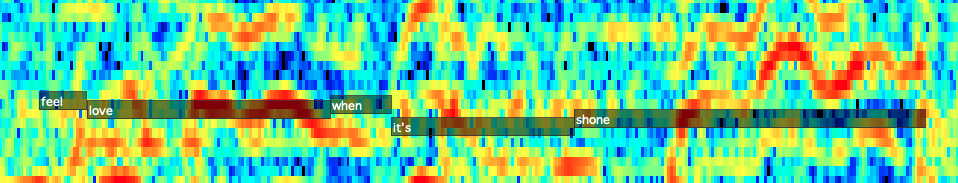}
    \caption{The CQT of an excerpt of the track in DALI with the lowest percentage error ($<1\%$ error), with it's annotations overlaid. The audio for this excerpt can be found at
    \url{https://youtu.be/Wq4tyDRhU_4?t=40}.}
    \label{fig:qual_ex}
\end{figure}

We manually inspected the tracks with a very high percentage of estimated errors ($> 70\%$) and found that all of them were the result of the annotation file being matched to the incorrect audio file (see~\cite{meseguerbrocal_2020} for details on the matching process).
On the other hand, we found that the tracks with a very low percentage of estimated errors ($< 1\%$) had qualitatively very high quality annotations.
For example, \figref{fig:qual_ex} shows an excerpt of the track with the lowest error rate along with a link to listen to the corresponding audio.
While this is only qualitative evidence, it is an additional indicator that the scores produced by the error detection model are meaningful.
The outputs of our model on DALI are made publicly available.

An error detection model that estimates the quality of a dataset can be used in several ways to improve the quality of a dataset.
For one, we can use it to direct manual annotation efforts both to the most problematic tracks in a dataset, but also to specific incorrect instances within a track.
Additionally, one of the major challenges regarding automatic annotation is knowing how well the automatic annotation is working.
For example, the creation of the DALI dataset used an automatic method for aligning the annotations with the audio, and it was very difficult for the creators to evaluate the quality of the annotations for different variations of the method.
This issue can now be overcome by using an error detection model to estimate the overall quality of the annotations for different variations of an automatic annotation method.

\section{Conclusions and Future Work}
In this paper we presented a novel data cleansing technique which considers the time-varying structure of labels.
We showed that it can be successfully applied to a dataset of vocal note event annotations, improving Raw Pitch Accuracy by over 10 percentage points simply by filtering the training dataset using our data cleansing model.
Our approach is particularly useful when training on very noisy datasets such as those collected from the Internet and automatically aligned.
We also used our proposed error detection model to estimate the error rate in the DALI dataset.

For future work, while our experiments focused on vocal note event annotations, we believe this technique could be directly applied to any kind of note event annotation, as well as extended for other types of time-varying annotations such as chords or beats.
We also believe the error detection model could be applied to scenarios other than training.
First, a natural use of such a model is to streamline manual annotation efforts by using the model to select time regions that are likely wrong and send them to an expert for correction.
Similarly, it could be used as an objective measure to guide the design of automatic annotation methods, which are otherwise forced to rely on manual evaluation.
We would also like to explore how this idea can be generalized to other domains beyond music and to test the contribution of different factors including the amount of noise in a dataset and the nature of the noise.

\section{Acknowledgements}
\thanks{Gabriel Meseguer-Brocal conducted this work at Spotify.}

\bibliography{main}

\end{document}